\documentclass{ceurart}
\usepackage{multirow}
\usepackage{tabularx}
\usepackage{makecell}
\usepackage{graphicx}
\usepackage{hyperref}

\usepackage{listings}

\begin{document}

\copyrightyear{2024}
\copyrightclause{Copyright for this paper by its authors.
  Use permitted under Creative Commons License Attribution 4.0
  International (CC BY 4.0).}

 \conference{}

\title{Unveiling Themes in Judicial Proceedings: A Cross-Country Study Using Topic Modeling on Legal Documents from India and the UK}

\author[1]{Krish Didwania}[%
orcid=0009-0007-1570-339X,
email=krishdidwania0674@gmail.com,
]

\address[1]{Department of Computer Science and Engineering, Manipal Institute of Technology, Manipal Academy of Higher Education, Manipal, India}

\author[2]{Durga Toshniwal}[%
email=durga.toshniwal@cs.iitr.ac.in
]
\address[2]{Professor, Department of Computer Science, Indian Institute of Technology, Roorkee, India}

\author[3]{Amit Agarwal}
\address[3]{Department of Computer Science, Indian Institute of Technology, Roorkee, India}

\footnotetext{Published at KM4LAW workshop at FOIS 2024}
\begin{abstract}
Legal documents are indispensable in every country for legal practices and serve as the primary source of information regarding previous cases and employed statutes. In today's world, with an increasing number of judicial cases, it is crucial to systematically categorize past cases into subgroups, which can then be utilized for upcoming cases and practices. Our primary focus in this endeavor was to annotate cases using topic modeling algorithms such as Latent Dirichlet Allocation, Non-Negative Matrix Factorization, and BerTopic for a collection of lengthy legal documents from India and the UK. This step is crucial for distinguishing the generated labels between the two countries, highlighting the differences in the types of cases that arise in each jurisdiction. Furthermore, an analysis of the timeline of cases from India was conducted to discern the evolution of dominant topics over the years.
\end{abstract}

\sloppy
\maketitle
\fussy

\section{Introduction}

A legal document holds paramount importance as a written testament, encapsulating contractual agreements, commitments, and legally binding actions. Renowned for their meticulous construction by legal experts, these documents ensure precision and accuracy \cite{jain2021summarization}. In this study, we delve into a collection of legal documents centered around court cases, capturing the intricate proceedings, decisions, and rulings within the judicial system. Serving as comprehensive records of legal disputes brought before courts, they document involved parties, issues, and judicial outcomes \cite{dodd1908modern}. Through these documents, one can trace the evolution of legal arguments, evidence presentation, and law application in addressing case complexities.

In today's era, abundant legal documents from accumulated judicial proceedings provide a vast data repository. The Supreme Court of India has witnessed significant developments in case disposal rates. In 2023, the apex court disposed of 52,191 cases, marking a 33\% increase compared to the previous year’s count of 39,800 cases. This achievement represents the highest disposal rate in the past six years. Our primary goal is to explore strategies leveraging this wealth of data to support future legal proceedings. Topic modeling emerges as a pivotal tool in this endeavor, automatically identifying underlying themes or topics within extensive document collections \cite{brookes2019utility}. By analyzing word distributions across documents, topic models eliminate the need for manual annotation, offering an efficient means to organize, explore, and index large datasets.

In our study, we employ topic modeling algorithms, including Latent Dirichlet Allocation (LDA) \cite{blei2003latent},Non-Negative Matrix Factorization(NMF)\cite{lee2000algorithms}, and BerTopic \cite{grootendorst2022bertopic}, to analyze legal documents from India and the UK. Beyond topic modeling, our research includes an ablation study examining judicial case types in both countries, comparing topic differences and semantic similarities. Additionally, we conduct a timeline analysis of Indian legal documents, observing trends in dominant topic changes over the years.

This research not only aims to understand the prevalent legal topics within these documents but also seeks to provide insights into the dynamics of legal proceedings and the evolving nature of legal discourse. Uncovering patterns and trends can enhance our understanding of legal systems and inform future legal practices and policies. Through this multidimensional analysis, we aim to contribute to the ongoing dialogue surrounding legal document analysis and its implications for the legal profession and society at large.

\section{Related Work}

Previous research underscores the significance of legal documents and their widespread implementation. Many of these studies focus on supervised learning techniques utilizing labels. Shukla et al.\cite{shukla2022legal} not only introduce the dataset used in our study but also provide summaries of judgments or segment-wise details, including facts, statutes, and analysis, through various supervised and unsupervised techniques. O et al.\cite{o2016analysis} concentrate on using topic models to summarize and visualize British legislation to facilitate easier browsing and identification of key legal topics and their associated terms.
Wang et al.\cite{wang2017topic} demonstrate the effectiveness and necessity of experiments to validate decision-making processes in the design, highlighting the high performance of the LDA algorithm in measuring text similarity. Similarly, Carter et al.\cite{carter2019proximity} conduct similar experiments on legal documents from the High Courts of Australia, focusing on case studies such as the Mabo litigation and the concept of 'proximity' in tort law.
Mohammadi et al.\cite{mohammadi2024combining} investigate the efficient handling of large-scale legal case law databases like Human Rights Documentation\cite{woods2017human}, particularly focusing on Article 8 of the European Convention on Human Rights, through topic modeling and citation networks. Kumar et al.\cite{kumar2012legal} propose an approach to generate concise summaries from legal judgments using topics obtained from LDA, providing a notable method for summarization, especially as the first such approach for Indian legal judgments.

Priyadarshini et al.\cite{priyadarshini2021ledocl} address instability in topic modeling through an ensemble approach, combining Semantic LDA and ensemble models, resulting in reduced processing time compared to conventional methods for legal texts from the UK. Regarding the variety of algorithms for topic modeling, Gonçalves et al.\cite{gonccales2015model} conduct a systematic mapping study to classify and analyze current literature, identifying trends and gaps in research areas and applied methods. Additionally, efforts have been made to enhance the learning of topic models by proposing regularization methods to improve coherence and interpretability, as suggested by Newman et al\cite{newman2011improving}.

While prior studies have examined legal texts from individual countries, our research, to the best of our knowledge, represents the first comparative study across multiple countries. Along with this, no previous work has incorporated a timeline analysis of legal documents from India.

\section{Methodolody}

\subsection{Dataset}

The dataset utilized in this study comprises three sections: Indian Abstractive, Indian Extractive, and UK Abstractive cases.

The Indian Abstractive dataset (IN-Abs) consists of Indian Supreme Court judgments obtained from the Legal Information Institute of India website, totaling 7,130 case documents with corresponding abstractive summaries.
These documents have an average token, 5389 in length.

The Indian Extractive dataset (IN-Ext) was curated based on feedback from legal experts dissatisfied with the IN-Abs summaries. Two LLB graduates annotated rhetorical segments in 50 Indian Supreme Court case documents and provided extractive summaries for each segment.

The UK Abstractive dataset (UK-Abs) comprises 793 case documents and their official press summaries from the UK Supreme Court website, segmented into abstractive summaries. These documents have average tokens, 14296 in length.

Notably, abstractive and extractive case summaries were not utilized in this paper as topic modeling employs unsupervised algorithms. During data examination, it was discovered that Indian cases spanned from 1945 to 2020, while UK cases only covered the years 2009 to 2010.

\subsection{Preprocessing}

In the preprocessing stage of topic modeling, we have employed some common practices to take various steps aimed at ensuring the accuracy of the analysis. These steps include the removal of stop words, which are commonly occurring words that contribute little semantic value and may distort the results. Additionally, lemmatization has been applied to standardize words by reducing them to their base or root form, thereby ensuring consistency among different inflections of the same word\cite{johnsen2019impact}.

During the implementation LDA and NMF, due to the extensive length of the documents, we eliminated frequently occurring common words found in judicial documents, especially those present in more than half of all cases. This process aimed to reduce the influence of ubiquitous terms during topic modeling, thus improving the distinctiveness and relevance of the resulting topics, which closely align with the underlying themes of the corpus. This preprocessing step significantly enhanced the quality of the outcomes. This procedure was not employed for BerTopic as better sentence embeddings would be generated for meaningful sentences.

\subsection{Topic Modelling}
This research work employed the following Topic Modelling algorithms for legal documents in both datasets:\\
\textbf{Latent Dirichlet Allocation (LDA):} 
LDA is a probabilistic model widely used for topic modeling in legal documents. It employs a two-step process: topic assignment and word generation.LDA utilizes the term frequency-inverse document frequency (TF-IDF)
\cite{christian2016single} top rioritize discriminative words in documents. In the context of our paper on legal document topic modeling, LDA acts as an unsupervised learning algorithm, extracting hidden topics by iteratively optimizing topic and word distributions to best explain the observed word occurrences. Overall, LDA offers a powerful approach for uncovering topics within legal documents, leveraging TF-IDF and probabilistic principles to capture their latent structure effectively.

\textbf{Non-Negative Matrix Factorization (NMF):} 
NMF is a dimensionality reduction technique widely used in topic modeling. In the context of legal documents, NMF decomposes the document-term matrix into two non-negative matrices: one representing topics and their distributions across words, and the other representing documents and their distributions across topics. This decomposition helps identify latent topics within the corpus of legal documents. Unlike LDA, NMF does not assume a probabilistic model but rather aims to factorize the input matrix into lower-dimensional matrices that capture meaningful patterns. In our application of NMF to legal document topic modeling, we utilized the TF-IDF vectorization technique in conjunction with NMF. TF-IDF is employed to transform the raw text data into a numerical representation that highlights the importance of words in individual documents relative to their occurrence across the entire corpus. The TF-IDF vectorization process assigns higher weights to words that are frequent within a document but relatively rare across the entire corpus, thereby emphasizing discriminative terms that are likely to be indicative of specific topics or themes. NMF is particularly suitable for legal document analysis as it ensures that all resulting factors are non-negative, which aligns well with the intuitive notion that topics and document-topic distributions should not contain negative values. By iteratively optimizing these matrices, NMF effectively extracts coherent topics that are interpretable in the context of legal terminology and concepts.

\textbf{BerTopic:} 
In our research, we utilize BerTopic, a topic modeling algorithm leveraging pre-trained BERT (Bidirectional Encoder Representations from Transformers) models to generate document embeddings from legal documents. These embeddings capture semantic meaning and are subsequently reduced in dimensionality using Uniform Manifold Approximation and Projection (UMAP)\cite{mcinnes2018umap}. UMAP preserves local and global structure, enabling efficient visualization and analysis of high-dimensional data. We employ MiniBatchKMeans clustering with 50 clusters to group similar documents, facilitating the identification of coherent topics. Preprocessing techniques, including OnlineCountVectorizer and ClassTfidfTransformer with BM25 weighting, enhance the quality and interpretability of resulting topics. To overcome the maximum input sequence limit of 512 for models like SentenceBert\cite{reimers2019sentence}, we segment input data into chunks, aggregating topics from different chunks for comprehensive topic extraction. This integration of UMAP with BerTopic enhances topic model interpretability and utility while efficiently handling BERT's input sequence limitation.

\section{Experimentation}

\subsection{Hyperparameter Tuning}

For LDA, hyperparameters are- the number of topics ($k$), $\alpha$ (parameter controlling the sparsity of document-topic distributions), and $\beta$ (parameter controlling the sparsity of topic-word distributions) were meticulously fine-tuned. We tested the model's performance using various combinations of hyperparameters, including different values of $\alpha$ and $\beta$ ranging from 0.01 to 0.99, both symmetric and asymmetric priors, and a range of $k$ values from 4 to 11. In the India dataset, optimal hyperparameters were determined as $\alpha = 0.46$, $\beta = 0.91$, and $k = 7$, while for the UK dataset, $\alpha$ was set to asymmetric, $\beta = 0.01$, and $k = 6$ \cite{panichella2021systematic}.

In implementing NMF, we opted for the same number of topics as LDA, as this algorithm requires less reliance on hyperparameter tuning. Furthermore, for BerTopic utilizing Sentence-BERT, specifying the number of topics beforehand is unnecessary. Instead, we adjusted other parameters related to dimensionality reduction and clustering to achieve optimal performance.

After determining the optimal $k$, resulting topics underwent expert annotation by legal law professionals.  Expert annotations served as a vital validation mechanism, refining the topic models by ensuring alignment with domain-specific nuances and requirements.

\subsection{Evaluation Metrics}
Topic modeling is a powerful technique used to extract underlying themes or topics from a collection of documents. However, assessing the quality of the topics generated by topic modeling algorithms is essential for ensuring their utility and interpretability. Coherence measures provide a quantitative assessment of the coherence and interpretability of topics by evaluating the semantic similarity between words within topics\cite{mimno2011optimizing}. It serves as a crucial metric for evaluating the quality of topics and assisting in model selection. 
In this work, we have evaluated all three models using two different coherence measures\cite{roder2015exploring}:
\\
a) C\_V coherence: This coherence measure calculates coherence based on the cosine similarity of word vectors. It evaluates the similarity between word pairs within topics by computing the cosine of the angle between their corresponding word vectors. The c\_v measure considers both the intra-topic coherence,i.e., similarity between words within a topic, and inter-topic coherence,i.e., similarity between words across different topics. 

\begin{equation}
C_v = \frac{1}{M} \sum_{i=1}^{M} \sum_{j=1}^{N_i - 1} \frac{{\text{similarity}(w_{i,j}, w_{i,j+1})}}{{\sqrt{{\sum_{k=1}^{N} w_{i,k}^2} \times \sum_{k=1}^{N} w_{i,(k+1)}^2}}}
\end{equation}
\\
\begin{equation}
\text{similarity}(\mathbf{w}_{i,j}, \mathbf{w}_{i,j+1}) = \frac{\mathbf{w}_{i,j} \cdot \mathbf{w}_{i,j+1}}{\|\mathbf{w}_{i,j}\| \cdot \|\mathbf{w}_{i,j+1}\|}
\end{equation}

\sloppy
Where \( C_v \) is the coherence score,\( M \) is the number of topics,\( N_i \) is the number of words in topic \( i \),\( w_{i,j} \) and \( w_{i,j+1} \) are two adjacent keywords in topic \( i \), \( \text{similarity}(w_{i,j}, w_{i,j+1}) \) is the word pair cosine similarity between \( w_{i,j} \) and \( w_{i,j+1} \).
\\
\fussy
b) U\_MASS: The u\_mass coherence measure quantifies coherence by measuring the pointwise mutual information (PMI) between pairs of words. It computes the PMI between all word pairs within topics and aggregates these scores to obtain the overall coherence score. The u\_mass measure assesses the semantic relatedness of words within topics based on their co-occurrence in the corpus.

It calculates how often two words, \(w_{i}\) and \(w_{j}\), appear together in the corpus and it’s defined as

\begin{equation} 
C_{UMass}(w_{i}, w_{j}) = \log \frac{D(w_{i}, w_{j}) + 1}{D(w_{i})}, 
\end{equation}

where \(D(w_{i}, w_{j})\) indicates how many times words \(w_{i}\) and \(w_{j}\) appear together in documents, and \(D(w_{i})\) is how many times word \(w_{i}\) appeared alone. The greater the number, the better the coherence score. Also, this measure isn’t symmetric, which means that \(C_{UMass}(w_{i}, w_{j})\) is not equal to \(C_{UMass}(w_{j}, w_{i})\). We calculate the global coherence of the topic as the average pairwise coherence scores on the top \(N\) words that describe the topic.

In the context of long document topic modeling, we have assigned utmost importance to the u\_mass coherence score. This metric holds significant weight as it evaluates the co-occurrence of keywords associated with topics throughout the entirety of long documents\cite{scheuter2021does}.

\section{Results}

\subsection{Quantitative Analysis}

In our statistical analysis, as displayed in Table \ref{tab:metrics}, we delved into the diverse array of topics present within legal documents from both India and the UK.

\begin{table}[ht]
\centering
\caption{Topic Coherence Scores for Different Models in India and UK}
\label{tab:metrics}
\begin{tabular}{cccc}
\hline
\multirow{2}{*}{Country} & \multirow{2}{*}{Model} & \multicolumn{2}{c}{Topic Coherence Scores} \\ \cline{3-4} 
                          &                         & \textbf{$c_v$}            & \textbf{$u_{\text{mass}}$}                    \\ \hline
\multirow{3}{*}{India}    & LDA                & \textbf{0.596}            & \textbf{-1.03}                      \\ 
                          & NMF                & \textbf{0.763}            & -1.915                  \\ 
            & BerTopic                & \textbf{0.781}            & -1.846                  \\ \hline
\multirow{3}{*}{UK}       & LDA                & \textbf{0.526}            & \textbf{-0.91}                    \\ 
                          & NMF                & \textbf{0.732}           & -0.915                      \\ 
                          & BerTopic                & \textbf{0.769}           & -1.554                   \\ \hline
\end{tabular}
\end{table}

\begin{figure}[ht]
  \centering
  \includegraphics[width=1\textwidth, height=0.27\textheight]{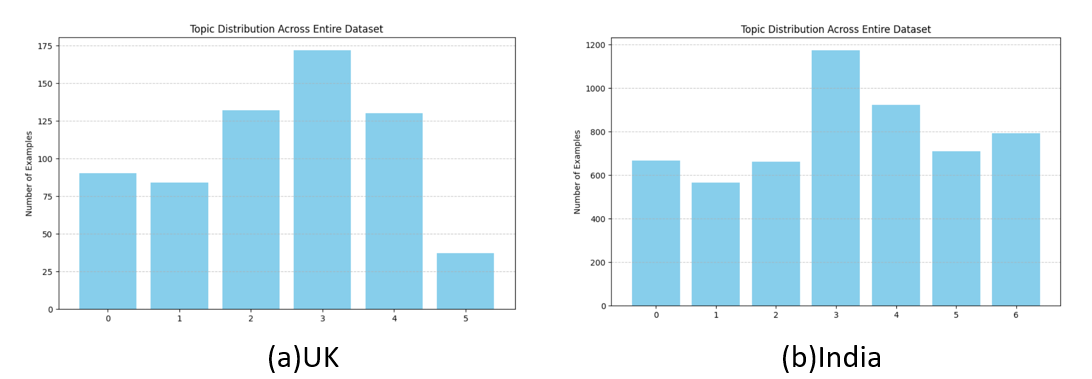} 
  \caption{Document frequency vs Topic ID Bar graph}
  \label{fig:dist}
\end{figure}

Notably, all three algorithms demonstrated semantic coherence within individual topics, showcasing the inherent similarity among words within each topic. Additionally, the analysis underscored the substantial diversity existing between different topics, indicating the richness and complexity of legal discourse. The bar graphs in Figure \ref{fig:dist} show the equal distribution of documents throughout the topics generated from the LDA model. This balanced distribution indicates that our model effectively mitigated the possibility of class imbalance, ensuring a comprehensive representation of various legal themes and issues within the dataset\cite{nikolenko2017topic}.

We observed that the LDA algorithm achieved the highest u\_mass score, highlighting its notable performance. This outcome shows the model's efficacy in capturing the underlying structure within lengthy legal texts.

\subsection{Comparison in topics of India and UK}

\sloppy
\begin{table}[ht]
\centering
\caption{Topics with Keywords and Annotations}
\label{tab:topics}
\begin{tabular}{ccp{0.4\textwidth}c}
\hline
\textbf{Country} & \textbf{Topic No.} & \textbf{Keywords} & \textbf{Annotation} \\
\hline
\multirow{7}{*}{\textbf{India}} 
& 1 & income, tax, service, assessee, rules, year, assessment, company, business, post & Income Tax \\
& 2 & election, candidate, public, petitioner, nomination, votes, religious, practice, corrupt, singh & Elections \\
& 3 & land, rights, legislature, public, lands, notification, area, clause, powers, parliament & Land Rights \\
& 4 & suit, property, decree, possession, rent, plaintiff, tenant, defendant, family, interest & Property Disputes \\
& 5 & tribunal, company, industrial, award, workmen, detention, dispute, board, employees, work & Industrial Disputes \\
& 6 & accused, code, offence, criminal, police, singh, magistrate, prosecution, pw, trial & Criminal Offences \\
& 7 & tax, goods, sale, sales, duty, contract, price, trade, excise, rules & Tax and Trade Regulations \\
\hline
\multirow{7}{*}{\textbf{UK}} 
& 1 & land, planning, development, notice, tenant, possession, tenancy, local, dismissal, property & Property \\
& 2 & directive, eu, member, arbitration, regulation, contract, tax, services, regulations, union & European Law\\
& 3 & criminal, convention, detention, police, trial, international, jurisdiction, offence, conviction, human & Human Rights \\
& 4 & child, children, tribunal, family, convention, life, care, immigration, treatment, parents & Family \\
& 5 & company, property, contract, value, tax, loss, jurisdiction, payment, clause, assets & Finance \\
& 6 & liability, injury, disease, insurance, employers, mesothelioma, caused, exposure, asbestos, employer & Insurance \\
\hline
\end{tabular}
\end{table}

\fussy

As demonstrated in Table \ref{tab:topics}, we collaborated with a legal expert to assign annotated labels based on the generated keywords for both datasets. These annotations also reveal differences in the predominant keywords between India and the UK.

\begin{figure}[ht]
  \centering
  \includegraphics[width=1\textwidth, height=0.2\textheight]{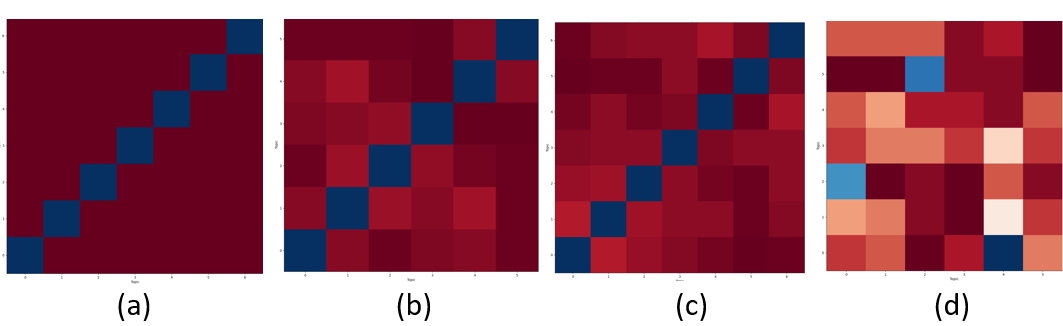} 
  \caption{Heatmaps depicting the interrelations among keywords across various topics within the LDA models: (a) Idealized topic correlations assuming no sequential relationship, (b) Topic correlations observed within the UK context, (c) Topic correlations evident in the India dataset, and (d) Comparative analysis of topic correlations between Indian and UK datasets.}
  \label{fig:heatmap}
\end{figure}

In the results section of legal documents focusing on topic modeling, it is worth noting the significant distinctions observed in the keywords extracted from legal texts originating from India and the UK \cite{alexander2015task}. This distinction emphasizes the notable diversity in the types of legal cases encountered in the two countries which sheds light on the uniqueness of their respective judicial systems and highlights the differences in the legal landscape, practices, and priorities between India and the UK \cite{agrawal2022judicial}.Such findings highlight the importance of considering regional and jurisdictional nuances when analyzing legal texts and stress the necessity for customized approaches in legal research and analysis.
The heatmaps in Figure \ref{fig:heatmap} (b) and (c) further confirm the diversity and discrepancy among the topics generated individually for both countries, while heatmap (d) illustrates the dissimilarity among the generated topics between India and the UK.

\subsection{Timeline Analysis of Indian cases}

\begin{figure}[ht]
  \centering
  \includegraphics[width=1\textwidth, height=0.25\textheight]{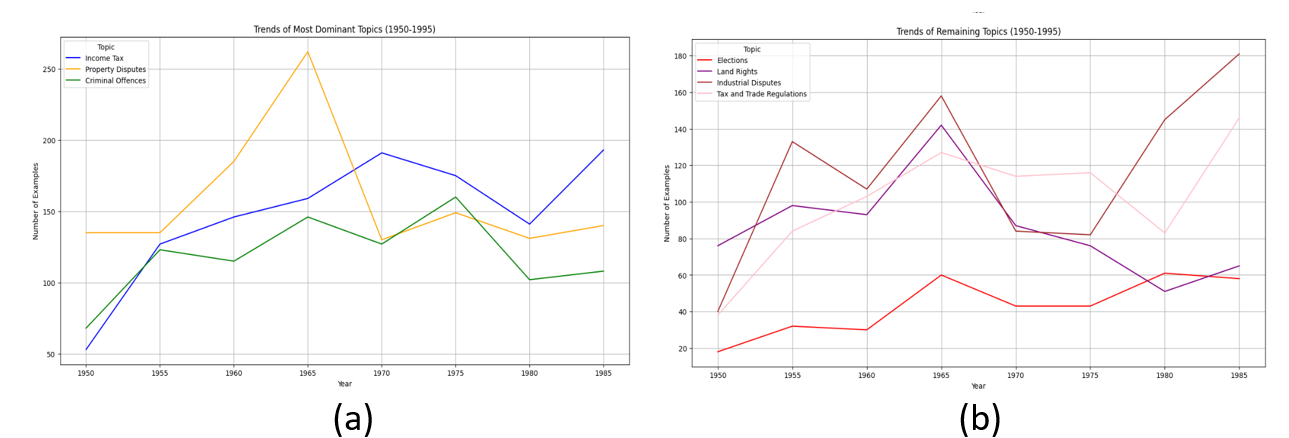} 
  \caption{Line graphs depicting the count of documents over the years for each topic: (a) Line graphs representing the three most prevalent topics, and (b) Line graphs illustrating the remaining four topics.}
  \label{fig:year}
\end{figure}

In this research, we also carried out a timeline analysis of Indian legal cases, examining the evolving trends in the primary subject matter over successive years\cite{linton2017dynamic}. The dataset covering Indian legal cases extended from 1945 to 2020, with the majority, constituting over 85\%, gathered between 1950 and 1990. We constructed a line graph as shown in Figure \ref{fig:year} illustrating document counts over this time frame, with each topic's involvement depicted distinctly.

The generated graphs portray the temporal dynamics of topic prevalence within the dataset spanning from 1950 to 1995. The graphs illustrate the trends for each topic, showcasing how their prominence fluctuated over the years. Each line represents a specific topic, and the y-axis indicates the number of documents pertaining to that topic. Both graphs together offer a comprehensive view of the thematic evolution within the corpus, shedding light on the shifts in focus and thematic trends across the specified time frame.

The surge in legal cases in India between 1950 and 1990 can be attributed to several intertwined factors\cite{ghosh1998economic}. Firstly, the era witnessed significant legislative reforms, possibly leading to confusion and disagreements that resulted in more disputes being brought to court. Rapid economic development spurred increased commercial activities, which in turn likely generated a higher number of legal conflicts over contracts, property rights, and taxation. Social and political changes, alongside a burgeoning population, may have further fueled civil unrest and disputes. Notably, the years 1955-1965 saw a peak in the number of cases, potentially influenced by the civil war at the time and the introduction of broad-based economic liberalization characterized by a blend of caprice, status quo-ism, and unfavorable economic conditions.

Despite the inconclusive nature of the line graphs, one can still discern notable trends. For instance, there is a marked upsurge in cases associated with income tax and trade regulations during those years, while topics such as land rights and criminal cases exhibit a significant decline after a specific period. The occurrences of industrial and property disputes and election cases fluctuated over time, experiencing periods of both surges and the absence of cases intermittently.

\section{Conclusion and Future Works}

In this study, we have utilized multiple topic modeling algorithms to analyze legal judicial cases from two countries: India and the UK. Within the legal domain, annotations of legal cases are imperative, serving as valuable resources for future cases and referencing past statutes. Our research demonstrates the effectiveness of employing topic modeling in automating annotation tasks, wherein generated keywords facilitate the identification of relevant topics. Furthermore, a noteworthy aspect of our study is the illustration of the disparities in the types of cases prevalent in both countries, thereby shedding light on variations in living standards and legal frameworks.

In our upcoming efforts within this project, we aspire to delve into the utilization of alternative transformer-based models and expansive language models, eliminating the necessity for segmentation. This approach will enhance the precision in identifying the specific topics relevant to each case. Moreover, our discoveries emphasize the vital need for a hierarchical framework in topic modeling. Such a structure could prove invaluable in scenarios requiring multi-label annotation, given that documents often relate to multiple subjects. Additionally, we intend to extend our timeline analysis to include more recent years post-1990, overcoming the limitations posed by the dataset's constraints and providing insights into evolving topic trends.
\fussy
\bibliography{sample-ceur}

\end{document}